\documentclass[runningheads]{llncs}
\usepackage{graphicx}
\usepackage{booktabs}
\usepackage{comment}
\usepackage{latexsym}
\usepackage{url}
\usepackage{graphicx}
\usepackage{xcolor}
\usepackage{multirow, multicol}
\usepackage{hyperref}
\usepackage{color,soul}
\usepackage{tcolorbox}

\usepackage[numbers,sort&compress]{natbib}

\begin{document}
\title{Multi-Task Learning for Features Extraction in Financial Annual Reports}
%
%

\author{Syrielle Montariol\inst{1} \and
Matej Martinc\inst{1} \and
Andraž Pelicon\inst{1} \and
Senja Pollak\inst{1} \and
Boshko Koloski\inst{1} \and
Igor Lončarski\inst{2} \and
Aljoša Valentinčič\inst{2} \and
Katarina Sitar Šuštar\inst{2}  \and
Riste Ichev\inst{2} \and
Martin Žnidaršič\inst{1} }
\authorrunning{Montariol et al.}
%
\institute{Jožef Stefan Institute, Jamova cesta 39, SI-1000 Ljubljana, Slovenia
\and
School of Economics and Business, University of Ljubljana, Kardeljeva pl. 17, SI-1000 Ljubljana, Slovenia
%
}

\maketitle              
\begin{tcolorbox}

 The final formatted version of this publication was published in  Proceedings of the MIDAS workshop of the Machine Learning and Principles and Practice of Knowledge Discovery in Databases  Conference (ECML-PKDD 2022), Grenoble, September, 2022 and is available online at \url{https://link.springer.com/chapter/10.1007/978-3-031-23633-4_1}.

\end{tcolorbox}
\begin{abstract}

For assessing various performance indicators of companies, the focus is shifting from strictly financial (quantitative) publicly disclosed information to qualitative (textual) information. This textual data can provide valuable weak signals, for example through stylistic features, which can complement the quantitative data on financial performance or on Environmental, Social and Governance (ESG) criteria.
 In this work, we use various multi-task learning methods for financial text classification with the focus on financial sentiment, objectivity, forward-looking sentence prediction and ESG-content detection. We propose different methods to combine the information extracted from training jointly on different tasks; our best-performing method highlights the positive effect of explicitly adding auxiliary task predictions as features for the final target task during the multi-task training. Next, we use these classifiers to extract textual features from annual reports of FTSE350 companies and investigate the link between ESG quantitative scores and these features. 

\keywords{Multi-task learning \and Financial reports \and Corporate Social Responsibility.}
\end{abstract}
\section{Introduction}

There is a slowly but steadily emerging consensus that qualitative (textual) information, when aiming at analysing a company's past, present and future performance, is equally if not more informative than quantitative (numerical) information.  Traditionally, financial experts and economists have used such qualitative information for decision making in a manual way. However, the volume of textual data increased tremendously in the past decades with the progressive dematerialisation and the growing capacity to share and store data \cite{LewisFinancialReports2019}, making the manual analysis practically infeasible.

Textual information about companies is found mainly in three contexts: mandatory public disclosures, news articles, and social media. Among all these data sources, periodic corporate reporting receives a particularly close attention from the research community, with an already plentiful literature in the financial domain and a growing one in natural language processing (NLP) \cite{chen2022overview, montariol-etal-2020-variations, masson2020detecting}. The reports are made publicly available periodically by all companies above a certain size and market value threshold, as defined by regulatory authorities of each country. The content of these periodical financial reports is also controlled by the regulators, with the goal of disclosing and communicating in great detail the financial situation and practices of the companies to investors \cite{sec1934}. 

Apart from the strictly quantitative financial information, these reports are rich in qualitative linguistic and structural information. This qualitative data can yield information about various financial aspects of companies at the time of filing, as well as predictions about future events in a company's financial ecosystem, such as future performance, stockholders' reactions, or analysts' forecasts \cite{lev1993fundamental,amir1996value}. Especially in sections that allow for a more narrative style, there is room for subjectivity and human influence in how financial data and prospects are conveyed.
Even though financial disclosures follow a reasonably well-established set of guidelines, there is still a great deal of variation in terms of how the content of these disclosures is expressed. The choice of specific words and tone when framing a disclosure can be indicative of the underlying facts about a company's financial situation that cannot be conveyed through financial indicators alone.

Extraction and processing of this information, however, prove to be much more challenging than for quantitative information.
There is a growing body of literature dedicated to the analysis of non-financial information in financial reports, where not only the content, but also stylistic properties of the text in reports are considered (e.g., \cite{Slattery2014, merkl2011impression}). 
For example, capturing and understanding the effect of information such as sentiment or subjectivity conveyed in the reports might give indications to predict investor behavior and the impact on supply and demand for financial assets \cite{formica2}.

Here, we study three key stylistic indicators associated with a text sequence: its sentiment, its objectivity, and its forward-looking nature. On top of this, we focus on a specific topic addressed by the annual reports: the Environmental, Social and Governance (ESG) aspects. These are part of the global Corporate Social Responsibility (CSR) framework, which refers to a set of activities and strategies that a company conducts in order to pursue pro-social and environmental objectives besides maximizing profit. Examples of these activities would include minimization of environmental externalities and charity donations. 

 As CSR is subject to a growing interest from investors, regulators and shareholders in the past few years,  companies have become more aware of how their actions and vocals impact the society and environment, prompting them to regularly report on their socio-environmental impact in the annual reports. The first requirements with regards to corporate social responsibility reporting were introduced by EU 2014/95 Directive (so called Non-Financial Reporting Directive or NFRD). \footnote{\url{https://ec.europa.eu/info/business-economy-euro/company-reporting-and-auditing/company-reporting/corporate-sustainability-reporting_en}} One of the proposed measures of CSR are the ESG criteria. The measure covers companies’ environmental impact (Environmental), relationships with their stakeholders --- e.g.  workplace conditions and impact of company's behaviour on the surrounding community --- (Social), and the integrity and diversity of its leadership (Governance), which considers criteria such as the accountability to shareholders and the accuracy and transparency of accounting methods. However, numerical indicators measuring the ESG performance of a company are few and far from being enough to evaluate precisely this concept. Similarly to other non-financial indicators, most ESG analyses are thus performed manually by experts \cite{lydenberg2010}. 

In this study, we aim at linking stylistic indicators with ESG-related concepts through multi-task learning, by fine-tuning pre-trained language models jointly on several classification tasks. More generally, we aim at proposing methods to extract features from textual reports using signals such as objectivity and forward-looking nature of sentences. 

Our contributions are as follows. We highlight the challenges of grasping the concepts of sentiment, objectivity and forward-looking in the context of finance by classifying content inside financial reports according to these categories. We compare several ways of exploiting these features jointly with the concept of ESG, through various multi-task learning systems. We show that we are able to make up for the difficulty of the tasks by training them in a multi-task setting, with a careful task selection. Moreover, we show that our ExGF multi-task learning system, where we Explicitly Give as Feature the predictions of auxiliary tasks for training a target task, beats other classical parameter sharing multi-task systems.
Finally, we provide qualitative insight into the link between ESG content, stylistic textual features and ESG numerical scores provided by press agencies.
Our code is available at \url{https://gitlab.com/smontariol/multi-task-esg}.

\section{Related Works}

\subsection{Annual Reports}

The literature on corporate Annual Reports (ARs) analysis is plentiful in the financial research community. From the NLP perspective, research is more scarce and much more recent. 
One of the most widely studied type of company reports are 10-K filings \cite{dyer2017evolution}. These are AR required by the U.S. Securities and Exchange Commission (SEC), and are so diligently studied thank to their format, which is highly standardised and controlled by the SEC. Outside the US, companies periodic reporting is less standardized and more shareholder-oriented. Here, we focus on UK annual reports. \citet{LewisFinancialReports2019} report significant increase in the size and complexity of UK annual report narratives: 
the median number of words more than doubled between 2003 and 2016\footnote{For a sample of 19,426 PDF annual reports published by 3252 firms listed on the London Stock Exchange.} while the median number of items in the table of contents also doubled in the same period. 
In face of these numbers, the automated analysis of financial reporting faces a growing contradiction: on the one hand, the huge increase in volume leads to the increased need for a solution from the NLP community to analyse this unstructured data automatically. On the other hand, more reporting from more companies leads to more diversity in the shape of the documents; this lack of standardization and structure makes the analysis tougher and requires more complex methods \cite{LewisFinancialReports2019}. 

In particular, concepts such as ESG are too recent to be included in any reporting standardization policy from the regulators, leading to very heterogeneous dedicated reporting. Consequently, as stated before, the work on detection of ESG-related content is somewhat scarce. Several works analyse the link between the publication of ESG report by a firm and its market value, without diving into the content of the report \cite{reverte2016corporate}.  \citet{formica1} use frequencies to capture semantic change of ESG-related keywords in UK companies' annual reports between 2012 and 2019. They interpret it using contextualised representations. A shared task aiming at classifying sentences as ``sustainable'' or not was organized in 2022 for the FinNLP workshop. \footnote{\url{https://sites.google.com/nlg.csie.ntu.edu.tw/finnlp-2022/shared-task-finsim4-esg}} The sentences were extracted from financial and non-financial companies' reports.

Another work closely related to our objective is the one from \citet{armbrust-etal-2020-computational}.
They underline the limited impact of the quantitative information in US companies annual reports, since detailed financial metrics and key performance indicators are often disclosed by the company before the publishing of the annual reports. Thus, as most of the financial information is redundant to the investors, regulators and shareholders, we turn towards stylistic features.



\subsection{Multi-Task Learning}

In this work, we investigate the methods to make use of various stylistic information to extract features from annual reports. We use pre-trained language models \cite{Devlin2019BERTPO} in a supervised multi-task learning (MTL) setting. The idea behind MTL, in which multiple learning tasks are solved in parallel or sequentially, relies on exploiting commonalities and differences across several related tasks in order to improve both training efficiency and performance of task-specific models.

The sequential MTL pre-training approach was formalized by \citet{phang2018sentence} under the denomination STILT (Supplementary Training on Intermediate Labeled-data Tasks), improving the efficiency of a pre-trained language model for a downstream task by proposing a preliminary fine-tuning on an intermediate task. This work is closely followed by \citet{pruksachatkun-etal-2020-intermediate}, who perform a survey of intermediate and target task pairs to analyze the usefulness of this intermediary fine-tuning. 
They find a low correlation between the acquisition of low-level skills and downstream task performance, while tasks that require complex reasoning and high-level semantic abilities, such as common-sense oriented tasks, had a higher benefit. 
\citet{aghajanyan2021muppet} extend the multi-task training to another level: they proposed ``pre-fine-tuning'', a large-scale learning step (50
tasks for around 5 million labeled examples) between language model pre-training and fine-tuning. They demonstrate that, when aiming at learning highly-generalizable representations for a wide range of tasks, the amount of tasks is key in multi-task training. Indeed, some tasks may hurt the overall performance of the system; but when performed in a large-scale fashion (e.g. with about 15 tasks), performance improved linearly with the number of tasks.

When it comes to parallel multi-task learning, the studies on this topic are abundant \cite{zhang2021survey}. While the research on the topic pre-dates the deep learning era, it has not been extensively researched until recently, when novel closely related transfer learning and pre-training paradigms \cite{Devlin2019BERTPO} proved very successful for a range of NLP problems. At the same time, multiple-task benchmark datasets, such as GLUE \cite{wang-etal-2018-glue} and the NLP Decathlon \cite{mccann2018natural} were released, offering new multi-task training opportunities and drastically reduced the effort to evaluate the trained models.
Due to these recent developments, several studies research weights sharing across networks and try to identify under which circumstances and for which downstream tasks MTL can significantly outperform single-task solutions \cite{worsham2020multi}. The study by \citet{standley2020tasks} suggested that MTL can be especially useful in a low resource scenario, where the problem of data sparsity can effect the generalization performance of the model. They claim that by learning to jointly solve related problems, the model can learn a more generalized internal representation. 
The crucial factor that determines the success of MTL is task similarity \cite{caruana1997multitask}. To a large extent, the benefit of MTL is directly correlated with the success of knowledge transfer between tasks, which improves with task relatedness. In cases when tasks are not related or only loosely related, MTL can result in inductive bias resulting in actually harming the performance of a classifier. 
Another factor influencing the success of the MTL performance is the neural architecture and design, e.g., the degree of parameters sharing between tasks \cite{ruder2017overview}. Here, we investigate several MTL architectures and methods and apply them to an annotated dataset of annual reports. 



\section{Datasets}

\subsection{FTSE350 Annual Reports and annotations}\label{sec:data_annot} 
The analysis is conducted on the same corpus as \cite{formica1}.\footnote{Code and details to re-create the dataset are available at \url{osf.io/rqgp4}.} It is composed of annual reports from the FTSE350 companies,
 covering the 2012-2019 time period. Only annual reports for London Stock Exchange companies that were listed on the FTSE350 list on 25th April 2020 are included. 
 Altogether 1,532 reports are collected in the PDF format and converted into raw text.

We use an annotated dataset associated with the FTSE350 corpus. The annotated sentences are extracted from reports covering the period between 2017 and 2019.
The annotators were given a sentence and asked to jointly label 5 tasks. In total, 2651 sentences were annotated. Here, we list the task definitions and label distributions.
\begin{itemize}
    \item \textbf{Relevance}: business related text (1768) / general text (883). Indicates whether the sentence is relevant from the perspective of corporate business.
    \item \textbf{Financial sentiment}: positive (1769) / neutral (717) / negative (165). Sentiment from the point of view of the financial domain.
    \item \textbf{Objectivity}: objective (2015) / subjective (636). Indicates whether the sentence expresses an opinion (subjectivity) or states the facts (objectivity).
    \item \textbf{Forward looking}: FLS (1561) / non-FLS (1090). Indicates whether the sentence is concerned with or planning for the future.
    \item \textbf{ESG}: ESG related (1889) / not ESG related (762). Indicates whether the sentence relates to sustainability issues or not.
\end{itemize}


\subsubsection{Inter-annotator agreement.}

In total, 13 annotators took part in the labeling process, all graduate students of MSc in Quantitative Finance and Actuarial Sciences. Among the annotated sentences, 48 sentences are annotated by all annotators. We use them to compute the inter-annotator agreement. To build the final corpus, we deduplicate the sentences used for inter-rater agreement, performing a majority vote to select the label for each sentence.

We compute three global measures of agreement between all annotators: Krippendorff's $\alpha$, Fleiss' $\kappa$, and the percentage of samples where all annotators agree. As a complement, we compute a pairwise measure: Cohen's $\kappa$ \cite{cohenkappa}, which we average between all pairs of annotators to obtain a global measure. It is similar to measuring the percentage of agreement, but taking into account the possibility of the agreement between two annotators to occur by chance for each annotated sample.
All these measures are indicated in Table \ref{tab:agreement}.


\begin{table}[!ht]
    \centering
\begin{tabular}{lc@{\hspace{5mm}}c@{\hspace{5mm}}c@{\hspace{5mm}}c}
\toprule
{} & Krippendorff $\alpha$ & Fleiss $\kappa$ & Cohen $\kappa$ &  Agreement (\%) \\
\midrule
Relevance &               0.09 &         0.09 &         0.10 &             6 \\
Financial sentiment  &               0.27 &         0.36 &        0.37 &            12 \\
Objectivity          &               0.26 &         0.26 &        0.25 &            18 \\
Forward looking               &               0.32 &         0.32 &        0.33 &            12 \\
ESG/not-Esg                &               0.43 &         0.42 &        0.43 &            28 \\
\bottomrule
\end{tabular}
\caption{Global measures of agreement between the 13 annotators on 48 sentences, for each of the 5 tasks.}
    \label{tab:agreement}
\end{table}

The agreement measures are consistently low. The tasks with the best inter-annotators agreements are sentiment and ESG. The Cohen $\kappa$ for ESG indicates a ``moderate'' agreement according to \cite{cohenkappa}, but it also indicates that less than one third of the annotations are reliable \cite{mchugh2012interrater}.
However, similar studies performing complex or subjective tasks such as sentiment analysis on short sentences also show low agreement values \cite{bobicev-sokolova-2017-inter}. 

\section{Multi-task Classification Methods}

Here, we tackle the classification tasks on the annotated dataset described in Section \ref{sec:data_annot}.
We use an encoder-decoder system for the classification tasks. A shared encoder is used to encode each sentence into a representation space, while different decoders are used to perform the classification. Here, we use the term `decoders' to denote the classification heads, which take as input the sentence representation encoded by the encoder part.

We describe the encoding and decoding systems in the following sections.

\subsection{Encoder: Pre-trained Masked Language Model}

Transformers-based pre-trained language models are a method to represent language, which establishes state-of-the-art results on a large variety of NLP tasks. Here, we use the RoBERTa model \cite{liu2019roberta}.
Its architecture is a multi-layer bidirectional Transformer encoder \cite{transformerVaswani}. The key element to this architecture, the bidirectional training, is enabled by the Masked Language Model training task: 15\% of the tokens in each input sequence are selected as training targets, of which 80\% are replaced with a [MASK] token. The model is trained to predict the original value of the training targets using the rest of the sequence. 

Contextualized language models are mostly used in the literature following the principle of transfer learning proposed by \citet{howard2018universal}, where the network is pre-trained as a language model on large corpora in order to learn general contextualised word representations. 

In our case, we perform domain adaptation \cite{peng-etal-2021-domain} by fine-tuning RoBERTa on the masked language model task on the FTSE350 corpus. Then, we perform a task-specific fine-tuning step for our various sentence classification tasks. To represent the sentences during the fine-tuning, we use the representation of the [CLS] token, which is the first token of every sequence. The final hidden state of the [CLS] token is usually used in sequence classification tasks as the aggregated sequence representation \cite{Devlin2019BERTPO}.


\subsection{Joint Multi-Task Learning} \label{sec:method_multi}

In the classical MTL setting, we implement a simple architecture taking as input each sentence's representation encoded by the contextualised language model, and feeding it to several decoders.
Each decoder is a classification head associated with one task. The sentence representation is passed through an architecture consisting of linear and dropout layers, before being projected to a representation space of the same dimension as the number of labels associated with the task. As a reminder, the financial sentiment task has three labels while the other ones have two. This final representation is called the \textit{logits}. We use them to compute the loss associated with each classification head.

To train all the decoders jointly with the encoder in an end-to-end fashion, we sum the losses outputted by each decoder, at each step. By optimizing this sum of losses, the model learns jointly on all tasks. In the results section, we denote this method as \textit{Joint}. We experiment with several task combinations to evaluate the positive and negative effects of each task on the performance of the classifier on other tasks.

\subsection{Weighting Tasks}\label{sec:method_multi_weighted}
Since the relatedness between tasks can vary, we also investigate an approach where weights for each task (i.e., the influence each task has on an average performance of the classifier across all tasks) are derived automatically. We train weights associated with each task jointly with training the classifiers, by adding an additional trainable ``weight layer'', containing $n$ scalars corresponding to $n$ tasks.
These weights are first normalized (i.e. the sum of weights is always one in order to prevent the model to reduce the loss by simply reducing all weights) and then multiplied to the calculated losses obtained for each task during training before summing up all losses and performing back-propagation. 

The learned weights can both be used to improve the overall performance, and to ``probe'' the importance of each task for the overall system.

\subsection{Sequential Multi-Task Learning} \label{sec:method_multi-seq}

Sequential MTL is an alternative to the joint MTL setting presented in Section \ref{sec:method_multi}. In the sequential setting, we use multi-task learning in an intermediary fashion, as a ``pre-fine-tuning'' for a given target task.
This is close to the concept of STILT \cite{phang2018sentence} and large-scale multi-task pre-fine-tuning presented in the related works section. The model, through a preliminary multi-task training step, is expected to acquire knowledge about all these training tasks and to accumulate it in the weights of the encoder. Then, the encoder is fine-tuned only on the downstream target task. 

In this setting, we experiment with various task combinations trained jointly before fine-tuning the encoder on one of the target tasks. We distinguish two sequential settings. First, systematically excluding the target task from the pre-fine-tuning step; for example, when the target task is the ESG classification, we pre-fine-tune the encoder using classification on various combinations of the four other tasks. Second, systematically including the target task; in that case, the encoder has seen the training data for the target task during pre-fine-tuning. In the results section, this method is referred to as \textit{Seq}.


\subsection{Explicitly Giving Features for Multi-Task Learning (ExGF - MTL)}

As explained before, task-specific classification heads output logits for each task. In this approach, we aim at performing multi-task learning by Explicitly Giving the output of the classification heads for ``auxiliary tasks'' as additional Features for the prediction of the final target task. These features are concatenated, fed into a linear layer (i.e., the ``common auxiliary task features' classification head''), and projected into a vector space of the same dimension as the logits of the final target task. The features are then summed with the logits of the ESG task, and this sum is used to compute the loss for the final target task. This final target task loss is then summed with the losses of the four auxiliary tasks (calculated on the logits outputted by the task-specific classification heads for each auxiliary task) and the backpropagation is performed the same way as for the joint MLT system.

\subsection{Task-aware Representations of Sentences (TARS)}

Task-aware representations \cite{halder2020task} were proposed as an alternative method to MTL where one general model can be trained and used for any task while preserving the maximum amount of information across tasks. The method transforms any classification task into a binary ``Yes/No'' classification task. Informally, for each task, a model is presented with a tuple consisting of a task label and an instance to classify. The task for the model to solve is to classify if the presented label matches with the presented instance or not. We adapt the proposed approach to model all five tasks jointly in a multitask setting. For a task with M classes, M $<$original label, instance$>$ tuples are generated for each instance in the training set. Each such generated instance is labeled with the generalized label ``Yes'' if the original label and instance match and with the generalized label ``No' otherwise. Original labels are prepended to the instance in order to condition the model both on the task and the instance to classify.

The model is trained jointly on instances for all tasks. Instances are generated in the same way during evaluation phase. As predicted class, we consider the class with the highest probability for the ``Yes'' generalized label. For example, given the sentence ``[...] colleagues were trained to deliver weekly walks targeted at individuals aged over 65.'' and the task of classifying if this sentence is ESG-related or not, the original label is ``ESG''. We create two instances and their respective labels:
\begin{itemize}
    \item ``ESG [SEP] \textit{sentence} ''. Gold label: ``Yes''. Probability of ``Yes'': 0.92.
    \item ``Not ESG [SEP] \textit{sentence} ''. Gold label: ``No''. Probability of ``Yes'': 0.37.
\end{itemize}
The model learns on all examples; at inference, given a similar pair of sentences and their ``Yes'' probability predicted by the model, we assign to the sentence the label associated with the highest probability of ``Yes''.

The motivation behind this system is related to the core idea behind multi-task training. All our systems implement a separate decoder for each task, as the task labels are different. Thus, the sharing of weights is only at the encoder-level. By gathering all tasks under a common setting, with the same binary labels, the setting is both more generalizable and fully shares all parameters during training. 
Recently, this idea of gathering all tasks under the same setting has gained a large popularity in NLP through the sequence-to-sequence paradigm \cite{aribandi2021ext5}.

\section{Multi-task Experiments}

In this section, we report the performance of the mono-task (used as a baseline) and the various multi-task settings introduced in the previous section.

\subsection{Experimental Framework}

In all experiments, the metric used to compare the systems is the macro-F1 score. It is the most adapted for classification settings with class imbalance (see Section \ref{sec:data_annot} for the label distribution for each task), as it computes performance for each label and averages them, without giving more importance to the over-represented ones. We divide the annotated dataset into 3 parts, keeping 20\% for development and 20\% for test. The test sets sampled from the original corpora are relatively small (531 examples). To increase the robustness of the results, we use five different seeds when fine-tuning the language model on the classification tasks and report the average scores over the five runs.

We tune hyper-parameters for the mono-task setting using the development set (batch size, learning rate, weight decay and number of epochs). The models are trained for 5 epochs and the selected model is the best one out of the 5 epochs in terms of average macro-F1 across all tasks (for joint multi-task learning) or best macro-F1 of the target task (sequential MTL or mono-task training), computed on the development set.


\subsection{Results}

Using each method for all task combinations, we compute the average macro-F1 score (across the five seeds) on the test set of each task. 
To compare the numerous methods and task combinations, we compute their rank in terms of performance (i.e. average macro-F1) for each target task. Then, we define the global performance of a method as the average rank of its performance across all target tasks, with a total number of methods and task combinations of 66. 
Table \ref{tab:ranked_best} shows the best-ranked methods and task combinations as well as a few lower-ranked methods for the sake of comparison. Task combinations are indicated by lists of integers; each integers corresponds to a task, and the matching is indicated in the caption. The N/As (Not Available) in the table correspond to the scores for a target task on which the system was not trained nor evaluated, because it is not part of the task combination.\footnote{Note that the N/As can only appear in the joint and weighted settings, where there is no explicit final target task.} The last column of the table shows the rank of the specific method and task combination among all approaches. For the sake of the further analysis and comparison between distinct architectures and task combinations, we include the last 4 lines of the table, which appear lower in the overall systems' ranking. 

With a relatively large margin, the best performing method is the ExGF-MTL system, leading to higher individual task scores compared with the mono-task training (in \textit{italic} in the table) for all tasks except financial sentiment. It is especially efficient on the ESG task. In terms of task combinations, the best systems exclude the Objectivity task (task $\#2$) and often the Relevance task (task $\#0$). These tasks have low inter-rater agreement, making them more difficult to tackle for the models. We further investigate their effect on MTL in the next table.

The last 4 lines of the table show systems with lower rankings. They allow us to compare the weighted MTL system trained on all task combinations, with the unweighted one; and the sequential and joint systems for the best task combinations. Except for the best task combination (i.e. 0-1-3-4, where sequential MTL beats joint MLT in terms of ranking (but by a low margin), the joint training seems overall better than the sequential one in the rankings. We further investigate this distinction in Table \ref{tab:mtl_res_part1}.

\begin{table}[!ht]
    \centering
\begin{tabular}{l@{\hspace{2mm}}lrrrr@{\hspace{8mm}}r@{\hspace{2mm}}r}
\toprule
         Method & Tasks & Relevance &  Fin-sentiment &  Objectivity &  Fwd-looking &   ESG &  rank \\
\midrule
ExGF & all &      51.67 &                58.67 &        \textbf{68.94} &            68.59 & \textbf{69.14 }&     \textbf{1} \\
 seq &  0-1-3-4 &      50.26 &                53.13 &        66.28 &            69.66 & 60.64 &     2 \\
    joint &   0-1-3-4 &      49.09 &                50.37 &          N/A &            \textbf{70.07 }& 64.07 &     3 \\
          \textit{mono} &    &  \textit{50.13} &                \textbf{\textit{63.11}} &        \textit{64.29} &           \textit{ 64.24 }& \textit{64.52} &     4 \\
weighted & all &      \textbf{54.70 }&                48.29 &        61.64 &            68.67 & 61.42 &     5 \\
     joint &    1-3-4 &        N/A &                50.94 &          N/A &            64.77 & 61.06 &     6 \\
         joint &         1-3 &        N/A &                49.56 &          N/A &            67.32 &   N/A &     7 \\ \midrule
     joint &  all &      48.80 &                48.34 &        63.53 &            66.00 & 55.31 &    19 \\
TARS & all &      49.48 &                46.42 &        59.62 &            66.49 & 59.86 &    24 \\
 seq &    1-3-4 &      48.30 &                47.43 &        59.20 &            60.13 & 52.69 &    39 \\
seq & all &      46.12 &                48.15 &        59.03 &            57.05 & 51.01 &    43 \\
\bottomrule
\end{tabular}

    \caption{Performance of all tested approaches according to the macro-F1 score arranged according to the average rank across tasks, i.e. the ranking of a row is the average of the five column-wise rankings, ignoring missing values (N/As). The task combinations are indicated as a list of integers with the following matching: 0 = Relevance; 1 =  Financial-sentiment; 2 =  Objectivity; 3 = Forward-looking; 4 = ESG. \textit{Mono} corresponds to the baseline classification results without multi-task learning.}
    \label{tab:ranked_best}
\end{table}

The last comparison is performed between weighted multi-task joint training (rank \#5) and non-weighted (rank \#19) training on all tasks; the joint method is outperformed by the weighted method on all tasks but one, the Objectivity task. The task importance according to the weights obtained during the weighted training are as follows: \{Relevance: -0.64; Fin-sentiment: 1.83;  Objectivity: -1.42; Fwd-looking: 1.6; ESG: -1.37\}.

For the sake of clarity, instead of showing the raw weights, we report the difference between the learned weights and the default uniform weights used in the other multi-task settings. Higher weights are given to the financial sentiment and the forward-looking tasks, while lower weights are given to the other three. Note that among the three tasks with the lower weights, two of them -- relevance and objectivity -- have the lowest inter-rater agreement (see Table \ref{tab:agreement}). 

To compare the various methods from a more global point of view, we average the scores of all task combinations for each method. Additionally, we also report on results for averages across all tasks but tasks \#0 (Relevance) and \#2 (Objectivity), the two tasks with the worst inter-annotator agreement, to further analyse the impact of removing low-performance tasks out of the MTL framework. The results are reported in Table \ref{tab:mtl_res_part1}\footnote{Note that the results for each method reported in Table \ref{tab:mtl_res_part1} are lower than the results reported in Table \ref{tab:ranked_best}, since here we report the average method's performance across all task combinations, while in Table \ref{tab:ranked_best} we only report results for the best ranked task combinations for each specific method.}. 

Removing the Objectivity and Relevance tasks leads to a small improvement in several cases, particularly for the joint training, where the weights are more directly impacted by the MTL system. The positive effect is less visible for the sequential training. It shows that for a small amount of tasks, or when a limited amount of data is available, more is often better.

The results also show that on average, regardless of the task combination, joint MTL tends to lead to better performance. Moreover, as expected, including the target task in the pre-fine-tuning step of the sequential training is highly beneficial, as it allows the model to learn interactions between this task and the auxiliary ones. However, note that this tactic is not in line with the initial idea behind large-scale pre-fine-tuning, which is to allow the model to be able to generalize to \textit{unseen} tasks (i.e. it is assumed that the target task data is not available during the preliminary step).

\begin{table}[!ht]
    \centering
\begin{tabular}{lcrrrr@{\hspace{8mm}}r}
\toprule

Method & Except  &  Relevance &  Fin-sentiment &  Objectivity &  Fwd-looking &    ESG \\\midrule
joint &                     &       47.9 &                48.78 &        57.38 &            66.18 &   56.80 \\
joint & 0             &        N/A &                48.56 &        60.32 &            67.41 &  57.82 \\
joint & 2             &      49.33 &                48.41 &          N/A &            67.17 &  54.88 \\
seq &          &      42.95 &                 40.7 &        57.23 &            55.37 &  49.88 \\
seq & 0 &      N/A &                38.59 &        56.78 &            54.57 &  49.44 \\
seq & 2 &      44.41 &                37.24 &        N/A &            55.95 &  51.35 \\
seq-INCL &           &      47.46 &                46.17 &        58.13 &            59.23 &  51.13 \\
\bottomrule
\end{tabular}
\caption{Macro-F1 average score over the 5 seeds and over all tasks combinations for the multi-task systems, \textit{except} some selected tasks when indicated in the second column. Seq-INCL means sequential training WITH the target task being already part of the pre-fine-tuning step.}
    \label{tab:mtl_res_part1}
\end{table}

\section{Linking ESG Ratings with Textual Features}

In the previous section, we identified ExGF-MTL as the best method to optimally exploit the information from all five tasks to improve the overall performance. We use this method to extract features from the annual reports, for all target tasks. Our aim is to investigate, on the large corpus of entire FTSE350 reports (i.e. not just on the manually labeled sentences extracted from these reports), the correlation between our extracted features and real-world numerical measures of ESG performance, obtained from Reuters, associated with each report.

\subsection{Inference on Reports}

\subsubsection{Pre-processing the corpus.}

We filter sentences in the corpus according to several conditions; these conditions are mainly related to the noise induced by the transformation from pdf to text, and to the corpus artefacts inherent to the format of the annual reports. We filter sentences depending on the proportion of upper-cased words and the proportion of non-letter characters in the sentence. We also filter short sentences not containing enough characters and words due to corpus artefacts such as the presence of split words into space-separated letters. Finally, we keep only sentences starting with a capital letter and ending with a punctuation sign. Following the pre-processing, we obtain an average number of around 1900 sentences by report.

\subsubsection{Feature Extraction.}

Using the ExGF-MTL models trained in the previous section, we perform inference on the full FTSE350 annual reports corpus. Thus, for each sentence, we predict its label associated to each task. We extract the following five features from each annual report: first, the proportion of ESG sentences. Then, among ESG sentences, the proportion of positive, negative, forward-looking, and objective sentences. Note that the Financial sentiment task is divided into two features to only get a set of binary features to compute the Spearman correlation.

\subsection{Correlation Analysis}


We use ESG scores from the financial press agency Reuters; for each company of the FTSE350 index and for each year, one score per ESG pillar (Environment, Governance and Social) and one Global ESG score are provided. They range from 0 to 100. They are inferred by financial analysts through careful reading of the financial reports of these companies.


We correlate these scores, using the Spearman correlation, with the five textual features extracted from the reports using the classification models. We also correlated the textual features with the year of the report, as an ordinal variable.
We perform the correlation analysis by grouping companies by ICB industry code (11 industries) and by extracting the 5 highest correlations between the two groups of features (numerical an text-extracted) in Table \ref{tab:qualitative}.

First, we note that the pillar score (Environment, Governance or Social) most correlated with the textual features is often related to the industry of the company (e.g. the Environment pillar for the Energy industry, the Governance pillar for the Financial industry). Among the most correlated text features with the ESG scores, the proportion of ESG sentences is often the highest, meaning that writing more about ESG in the reports is often linked with having good ESG scores. Following closely in terms of Spearman correlation, is the percentage of negative and objective ESG sentences. We also note that the year is often correlated with the proportion of ESG sentences, meaning that ESG is increasingly discussed in the recent years. The proportion of forward-looking sentences is seldom correlated with the ESG scores; a notable correlation is the one between the forward-looking proportion and Governance Pillar score in the financial industry, which indicates a specific writing style about governance in this industry. 

\begin{table}[!ht]
    \centering
 \resizebox{0.48\textwidth}{!}{
 \begin{tabular}{lllr}
\toprule
Industry & \begin{tabular}[x]{@{}c@{}}Reuters \\ Scores (\%)\end{tabular} & \begin{tabular}[x]{@{}c@{}}Textual \\ Features (\%)\end{tabular} & $\rho$ \\
\midrule
\multirow{5}{*}{Energy} & Environment & Objective & 0.49 \\
& Social & Objective & 0.45 \\
& Environment & ESG & 0.43 \\
& Social & ESG & 0.42 \\
& Global & Objective & 0.42 \\

\midrule

\multirow{5}{*}{\begin{tabular}[x]{@{}c@{}}Consumer \\ Staples\end{tabular}} & Global & ESG & 0.43 \\
& Social & ESG & 0.42 \\
& Governance & ESG & 0.33 \\
& Environment & ESG & 0.33 \\
& Environment & Negative & 0.30 \\

\midrule

\multirow{5}{*}{Industrials}  & Social & Objective & 0.41 \\
& Global & Objective & 0.38 \\
& Environment & Objective & 0.32 \\
& year & Objective & 0.26 \\
& year & ESG & 0.17 \\

\midrule

\multirow{5}{*}{\begin{tabular}[x]{@{}c@{}}Telecom-\\munication\end{tabular}} & year & ESG & 0.44 \\
& Environment & ESG & 0.41 \\
& year & Positive & 0.41 \\
& year & Objective & 0.38 \\
& Environment & Fwd-looking & 0.32 \\

\midrule

\multirow{5}{*}{\begin{tabular}[x]{@{}c@{}}Real \\ Estate\end{tabular}} & year & ESG & 0.32 \\
& Environment & Negative & 0.28 \\
& Social & ESG & 0.26 \\
& year & Objective & 0.23 \\
& Global & ESG & 0.23 \\

\midrule

\multirow{2}{*}{\begin{tabular}[x]{@{}c@{}}Basic \\ Materials\end{tabular}} & Governance & Fwd-looking & 0.22 \\
& year & Objective & 0.18 \\

\midrule
\end{tabular}

}
 \resizebox{0.48\textwidth}{!}{
 \begin{tabular}{lllr}
\toprule
Industry & \begin{tabular}[x]{@{}c@{}}Reuters \\ Scores (\%)\end{tabular}  & \begin{tabular}[x]{@{}c@{}}Textual \\ Features (\%)\end{tabular} & $\rho$ \\
\midrule
\multirow{5}{*}{Utilities}& Social & ESG & 0.55 \\
& Global & Negative & 0.39 \\
& Social & Objective & 0.33 \\
& Governance & Negative & 0.31 \\
& year & ESG & 0.29 \\

\midrule

\multirow{5}{*}{Financials} & Governance & Fwd-looking & 0.29 \\
& Governance & Negative & 0.28 \\
& Governance & ESG & 0.25 \\
& Global & Objective & 0.24 \\
& year & Objective & 0.22 \\

\midrule

\multirow{5}{*}{\begin{tabular}[x]{@{}c@{}}Consumer \\ Discretionary\end{tabular}} & Environment & Negative & 0.31 \\
& Environment & Objective & 0.30 \\
& Global & Objective & 0.26 \\
& year & Objective & 0.21 \\
& Social & Fwd-looking & 0.20 \\

\midrule

\multirow{5}{*}{Technology} & Governance & ESG & 0.54 \\
& Environment & Objective & 0.42 \\
& Global & ESG & 0.39 \\
& Governance & Negative & 0.35 \\
& Social & Negative & 0.34 \\

\midrule

\multirow{5}{*}{\begin{tabular}[x]{@{}c@{}}Health \\ Care\end{tabular} } & Social & Objective & 0.29 \\
& year & Positive & 0.29 \\
& year & Objective & 0.25 \\
& Global & Objective & 0.24 \\
& Governance & Objective & 0.20 \\

\bottomrule
\end{tabular}
}
    \caption{Top 5 highest pairwise Spearman correlations between textual features (proportion of positive, negative, objective ESG sentences \& proportion of ESG sentences in the full report), Reuters ESG scores (Global, Env., Social and Gov.) and year, by industry. Only Spearman correlations higher than 0.18 are displayed.}
    \label{tab:qualitative}
\end{table}

\section{Conclusion}

The focus of this study was the joint use of stylistic features -- financial sentiment, objectivity and forward-looking -- and ESG classification.
We turned towards ESG because it is a challenging concept to quantify and evaluate, partly because few numerical metrics exist to characterize it, and because it is a recent concept that is still not perfectly defined and structured. But our experimental framework and MTL methods are generic and can be applied to any other topic of interest, for investigating the correlation between financial and textual data.

Methodologically, we showed that the best way to combine information from related tasks is to explicitly provide the predictions of auxiliary tasks as features for the prediction of a target task. This system is even efficient for tasks made very challenging by a very low inter-rater agreement. 
Note that in the majority of the literature, MTL is performed using different datasets for each task, while in our case, all the tasks are included in a single dataset, each instance having a label for each task. However, the proposed best approach, ExGF-MTL, can easily be applied to the multi-dataset multi-task learning; the only difference being the data loading implementation and encoder-decoder interactions. 
We also showed the importance of task selection when performing multi-task learning with a low number of tasks. We posit that a higher number of tasks would allow the system to compensate for low-performance tasks. When identifying low-performance tasks that harm the MTL system, we highlighted the link between performance and annotation quality. 
Finally, following recent trends for large-scale NLP multi-task learning, we compared sequential and joint fine-tuning, and experimented with MTL using a unique decoder for all tasks (TARS). However, we could not show any positive effect of this latter method. A better way to use a unique decoder for MTL while making the most out of large pre-trained language models would be to adopt the sequence-to-sequence paradigm for all tasks \cite{aribandi2021ext5}.

Qualitatively, we showed that our method allows us to extract meaningful features from annual reports that correlate with numerical features provided by press agencies, on the topic on corporate social responsibility. In future work, we plan to extend our method to perform causal discovery and causal inference between textual features, ESG scores and various financial performance indicators for companies \cite{keith-etal-2020-text}. 

\section*{Acknowledgements}

This work was supported by the Slovenian Research Agency (ARRS) grants for the core programme Knowledge technologies (P2-0103) and the project quantitative and qualitative analysis of the unregulated corporate financial reporting (J5-2554). We also want to thank the students of the SBE for their effort in data annotation.

%
%
%
\bibliographystyle{splncs04nat}
\bibliography{biblio.bib}

\end{document}